\documentclass[sigconf]{acmart}

\usepackage{subfig}
\usepackage{todonotes}

\AtBeginDocument{%
  \providecommand\BibTeX{{%
    \normalfont B\kern-0.5em{\scshape i\kern-0.25em b}\kern-0.8em\TeX}}}

\copyrightyear{2020} 
\acmYear{2020} 
\setcopyright{rightsretained} 
\acmConference[CIKM '20]{Proceedings of the 29th ACM International Conference on Information and Knowledge Management}{October 19--23, 2020}{Virtual Event, Ireland}
\acmBooktitle{Proceedings of the 29th ACM International Conference on Information and Knowledge Management (CIKM '20), October 19--23, 2020, Virtual Event, Ireland}\acmDOI{10.1145/3340531.3412777}
\acmISBN{978-1-4503-6859-9/20/10}
\begin{document}

\title{\texttt{Falcon 2.0}: An Entity and Relation Linking Tool over Wikidata}

\author{Ahmad Sakor}
\email{ahmad.sakor@tib.eu}
\affiliation{%
  \institution{L3S Research Center and TIB, University of Hannover}
  \city{Hannover}
  \country{Germany}
}

\author{Kuldeep Singh}
\email{kuldeep.singh1@cerence.com}
\affiliation{%
  \institution{Cerence GmbH and Zerotha Research}
  \streetaddress{P.O. Box 1212}
  \city{Aachen}
  \country{Germany}
}

\author{Anery Patel}
\email{anery.patel@tib.eu}
\affiliation{%
  \institution{TIB, University of Hannover}
  \city{Hannover}
  \country{Germany}}

\author{Maria-Esther Vidal}
\email{maria.vidal@tib.eu}
\affiliation{%
  \institution{L3S Research Center and TIB, University of Hannover}
  \city{Hannover}
  \country{Germany}}

\renewcommand{\shortauthors}{Sakor, et al.}

\begin{abstract}
The Natural Language Processing (NLP) community has significantly contributed to the solutions for entity and relation recognition from a natural language text, and possibly linking them to proper matches in Knowledge Graphs (KGs). Considering Wikidata as the background KG, there are still limited tools to link knowledge within the text to Wikidata.
In this paper, we present \texttt{Falcon 2.0}, the first joint entity and relation linking tool over Wikidata. It receives a short natural language text in the English language and outputs a ranked list of entities and relations annotated with the proper candidates in Wikidata.
The candidates are represented by their Internationalized Resource Identifier (IRI) in Wikidata. \texttt{Falcon 2.0} resorts to the English language model for the recognition task (e.g., N-Gram tiling and N-Gram splitting), and then an optimization approach for the linking task. We have empirically studied the performance of \texttt{Falcon 2.0} on Wikidata and concluded that it outperforms all the existing baselines. \texttt{Falcon 2.0} is open source and can be reused by the community; all the required instructions of  \texttt{Falcon 2.0} are well-documented at our GitHub repository\footnote{\url{https://github.com/SDM-TIB/falcon2.0}}. We also demonstrate an online API, which can be run without any technical expertise.
\texttt{Falcon 2.0} and its background knowledge bases are available as resources at \url{https://labs.tib.eu/falcon/falcon2/}.

\end{abstract}




\maketitle

\section{Introduction}
Entity Linking (EL)- also known as Named Entity Disambiguation (NED)- is a well-studied research domain for aligning unstructured text to its structured mentions in various knowledge repositories (e.g., Wikipedia, DBpedia~\cite{DBLP:conf/semweb/AuerBKLCI07}, Freebase~\cite{DBLP:conf/aaai/BollackerCT07} or Wikidata~\cite{DBLP:conf/www/Vrandecic12}). Entity linking comprises two sub-tasks. The first task is Named Entity Recognition (NER), in which an approach aims to identify entity labels (or surface forms) in an input sentence. Entity disambiguation is the second sub-task of linking entity surface forms to semi-structured knowledge repositories. 
With the growing popularity of publicly available knowledge graphs (KGs), researchers have developed several approaches and tools for EL task over KGs. Some of these approaches implicitly perform NER and directly provide mentions of entity surface forms in the sentences to the KG (often referred to as end-to-end EL approaches) \cite{delpeuch2019opentapioca}. Other attempts (e.g., Yamanda et al.~\cite{DBLP:conf/conll/YamadaS0T16}, DCA~\cite{DBLP:conf/emnlp/YangGLTZWCHR19}) consider recognized surface forms of the entities as additional inputs besides the input sentence to perform entity linking. Irrespective of the input format and underlying technologies, the majority of the existing attempts ~\cite{roder2018gerbil} in the EL research are confined to well-structured KGs such as DBpedia or Freebase\footnote{it is now depreciated and no further updates are possible}. These KGs rely on a well-defined process to extract information directly from Wikipedia infoboxes. They do not provide direct access to the users to add/delete the entities or alter the KG facts. Wikidata, on the other hand, also allows users to edit Wikidata pages directly, add newer entities, and define new relations between the objects. Wikidata is hugely popular as a crowdsourced collection of knowledge. Since its launch in 2012, over 1 billion edits have been made by the users across the world\footnote{\url{https://www.wikidata.org/wiki/Wikidata:Statistics}}. 

\paragraph{Motivation, Approach, and Contributions.} 
We motivate our work by the fact that despite the vast popularity of Wikidata, there are limited attempts to target entity and relation linking over Wikidata. For instance, there are over 20 entity linking tools/APIs for DBpedia \cite{DBLP:conf/www/SinghRBSLUVKP0V18,roder2018gerbil}, which are available as APIs. To the best of our knowledge, there exists only one open-source API for Wikidata entity linking (i.e., OpenTapioca~\cite{delpeuch2019opentapioca}). Furthermore, there is no tool over Wikidata for relation linking, i.e., linking predicate surface forms to their corresponding Wikidata mentions. In this paper, we focus on providing \texttt{Falcon 2.0}, a reusable resource API for joint entity and relation linking over Wikidata. In our previous work, we proposed Falcon~\cite{sakor2019old}, a rule-based approach yet effective for entity and relation linking on short text (questions in this case) over DBpedia. In general, the Falcon approach has two novel concepts: 1) a linguistic-based approach that relies on several English morphology principles such as tokenization, and N-gram tiling; 2) a local knowledge base which serves as a source of background knowledge (BK). This knowledge base is a collection of entities from DBpedia.
We \textbf{resort} to the Falcon approach for developing \texttt{Falcon 2.0}. Our aim here is to study whether or not the Falcon approach is agnostic to underlying KG; hence, we do not claim novelty in the underlying \emph{linguistic-based approach} for \texttt{Falcon 2.0}. Further, we investigate the concerns related to robustness, emerging failures, and bottlenecks. We introduce \texttt{Falcon 2.0} based on the methodology employed in the first version. Our tool is the \textit{first} joint entity and relation linking tool for Wikidata. Our novel contributions briefly lie in two aspects: 

\begin{enumerate}
\item \textbf{\texttt{Falcon 2.0}}: 
   The first resource for joint entity and relation linking over Wikidata. \texttt{Falcon 2.0} relies on fundamental principles of English morphology (tokenization and compounding) and links entity and relation surface forms in a short sentence to its Wikidata mentions. \texttt{Falcon 2.0} is available as an online API and can be accessed at \url{https://labs.tib.eu/falcon/falcon2/}. Falcon 2.0 is also able to recognize entities in keywords such as Barack Obama, where there is no relation. We empirically evaluate \texttt{Falcon 2.0} on three datasets tailored for Wikidata. According to the observed results, \texttt{Falcon 2.0} significantly outperforms all the existing baselines. For the ease of use, we integrate the Falcon API\footnote{\url{https://labs.tib.eu/falcon/}} into Falcon 2.0.
   This option is available in case Wikipedia contains an equivalence entity (Wikidata is a superset of DBpedia)
   The \texttt{Falcon 2.0} API already has over half a million hits from February 2020 to the time of paper acceptance, which shows its gaining usability (excluding self-access of the API while performing the evaluation). 
    
\item \textbf{\texttt{Falcon 2.0 Background KG}}: 
    We created a new background KG of \texttt{Falcon 2.0} with the Wikidata. We extracted 48,042,867 Wikidata entities from its public dump and aligned these entities with the aliases present in Wikidata. For example, Barack Obama is a Wikidata entity Wiki:Q76\footnote{\url{https://www.wikidata.org/wiki/Q76}}. We created a mapping between the label (Barack Obama) of Wiki:Q76 with its aliases such as President Obama, Barack Hussein Obama, and Barry Obama and stored it in the background knowledge base. We implemented a similar alignment for 15,645 properties/relations of Wikidata. The background knowledge base is an indexed graph and can be queried. The resource is also present at a persistent URI for further reuse\footnote{\url{https://doi.org/10.6084/m9.figshare.11362883}}.
    
    
\end{enumerate}

The rest of this paper is organized as follows: Section \ref{sec:related} reviews the state-of-the-art, and the following Section \ref{sec:falcon} describes our two resources and approach to build \texttt{Falcon 2.0}. Section \ref{sec:experiment} presents experiments to evaluate the performance of \texttt{Falcon 2.0}. Section \ref{sec:impact} presents the importance and impact of this work for the research community. The availability and sustainability of resources is explained in Section \ref{sec:adaptation}, and its maintenance related discussion is presented in Section~\ref{sec:maintance}. We close with the conclusion in Section~\ref{sec:conclude}.

\section{Related Work}
\label{sec:related}
Several surveys provide a detailed overview of the advancements of the techniques employed in entity linking over KGs~\cite{shen2015,balog_2018}. Various reading lists~\cite{hengji2019}, online forums\footnote{\url{http://nlpprogress.com/english/entity_linking.html}} and Github repositories\footnote{\url{https://github.com/sebastianruder/NLP-progress/blob/master/english/entity_linking.md}} track the progress in the domain of entity linking. Initial attempts in EL considered Wikipedia as an underlying knowledge source. The research field has matured and the SOTA nearly matches human-level performance~\cite{raiman2018deeptype}. With the advent of publicly available KGs such as DBpedia, Yago, and Freebase, the focus has shifted towards developing EL over knowledge graphs. The developments in Deep Learning have introduced a range of models that carry out both NER and NED as a single end-to-end step~\cite{kolitsas2018end,DBLP:conf/emnlp/GaneaH17}. NCEL~\cite{CaoYixin-2018} learns both local and global features from Wikipedia articles, hyperlinks, and entity links to derive joint embeddings of words and entities. These embeddings are used to train a deep Graph Convolutional Network (GCN) that integrates all the features through a Multi-layer Perceptron. The output is passed through a Sub-Graph Convolution Network, which finally resorts to a fully connected decoder. The decoder maps the output states to linked entities. The BI-LSTM+CRF model~\cite{Emrah-W18-2403} formulates entity linking as a sequence learning task in which the entity mentions are a sequence whose length equals the series of the output entities. Albeit precise, deep learning approaches demand \emph{high-quality} training annotations, which are not extensively available for Wikidata entity linking \cite{cetoli2019neural,mulang2019context}.

There is concrete evidence in the literature that the machine learning-based models trained over generic datasets such as  WikiDisamb30~\cite{DBLP:conf/cikm/FerraginaS10}, and CoNLL (YAGO)~\cite{DBLP:conf/emnlp/HoffartYBFPSTTW11} do not perform well when applied to short texts. Singh et al.~\cite{DBLP:conf/www/SinghRBSLUVKP0V18} evaluated more than 20 entity linking tools over DBpedia for short text (e.g., questions) and concluded that issues like capitalization of surface forms, implicit entities, and multi-word entities affect the performance of EL tools in a short input text. Sakor et al. ~\cite{sakor2019old} addresses specific challenges of short texts by applying a rule-based approach for EL over DBpedia. In addition to linking entities to DBpedia, Sakor et al. also provides DBpedia IRIs of the relations in a short text. EARL~\cite{banerjeejoint} is another tool that proposes a traveling salesman algorithm-based approach for joint entity and relation linking over DBpedia. To the best of our knowledge, EARL and Falcon are the only available tools that provide both entity and relation linking.

Entity linking over Wikidata is a relatively new domain. Cetoli et al.~\cite{cetoli2019neural} propose a neural network-based approach for linking entities to Wikidata. The authors also align an existing Wikipedia corpus-based dataset to Wikidata. However, this work only targets entity disambiguation and assumes that the entities are already recognized in the sentences. Arjun~\cite{mulang2019context} is the latest work for Wikidata entity linking. It uses an attention-based neural network for linking Wikidata entity labels. OpenTapioca~\cite{delpeuch2019opentapioca} is another attempt that performs end-to-end entity linking over Wikidata; it is the closest to our work even though OpenTapioca does not provide Wikidata Ids of relations in a sentence. OpenTapioca is also available as an API and is utilized as our baseline. S-Mart~\cite{yang2015s} is a tree-based structured learning framework based on multiple additive regression trees for linking entities in a tweet. The model was later adapted for linking entities in the questions. VCG~\cite{sorokin2018mixing} is another attempt which is a unifying network that models contexts of variable granularity to extract features for an end to end entity linking. 
However, \texttt{Falcon 2.0} is the \textit{first} tool for joint entity and relation linking over Wikidata.

\section{\texttt{Falcon 2.0}- A Resource}\label{sec:falcon}
In this section, we describe \texttt{Falcon 2.0} in detail. First the architecture of \texttt{Falcon 2.0} is depicted. Next, we discuss the BK used to match the surface forms in the text to the resource in a specific KG. In the paper's scope, we define "short text" as grammatically correct questions (up to 15 words).

\subsection{Architecture}
\begin{figure*}[t]
	\centering
	\includegraphics[width=1\textwidth]{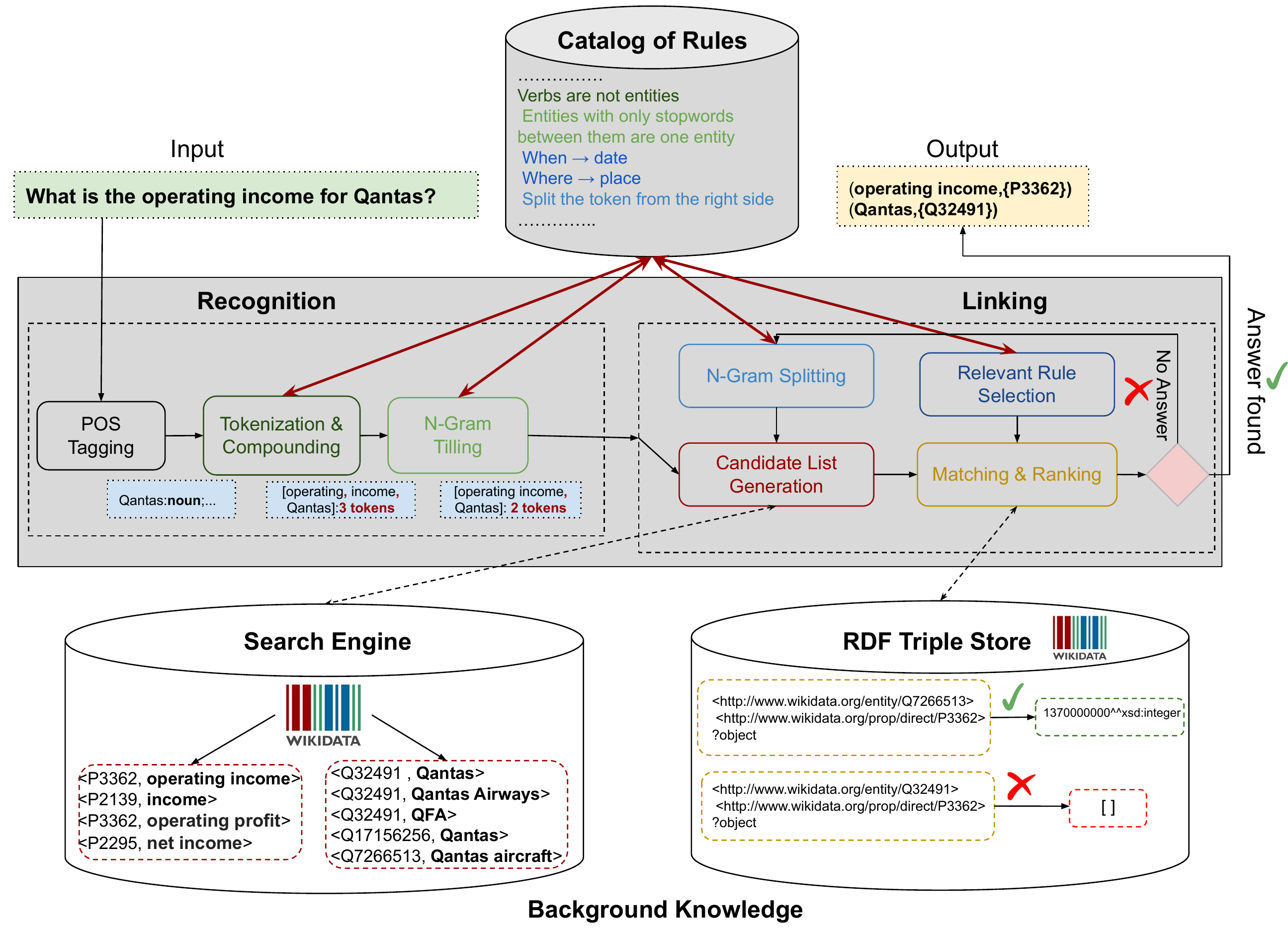}
	\caption{The \textbf{\texttt{Falcon 2.0} Architecture}. The boxes highlighted in Grey are reused from Falcon \cite{sakor2019old}. Grey boxes contain a linguistic pipeline for recognizing and linking entity and relation surface forms. The boxes in White are our addition to the Falcon pipeline to build a resource for the Wikidata entity and relation linking. The white boxes constitute what we refer to as BK specific to Wikidata. The text search engine contains the alignment of Wikidata entity/relation labels along with the entity and relation aliases. It is used for generating potential candidates for entity and relation linking. RDF triple store is a local copy of Wikidata triples containing all entities and predicates.
	}
	\label{fig:archi}
\end{figure*}
The \texttt{Falcon 2.0} architecture is depicted in Figure \ref{fig:archi}. \texttt{Falcon 2.0} receives short input texts and outputs a set of entities and relations extracted from the text; each entity and relation in the output is associated with a unique Internationalized Resource Identifier (IRI) in Wikidata. \texttt{Falcon 2.0} resorts to BK and a catalog of rules for performing entity and relation linking. The BK combines Wikidata labels and their corresponding aliases. Additionally, it comprises alignments between nouns and entities in Wikidata. Alignments are stored in a text search engine, while the knowledge source is maintained in an RDF triple store accessible via a SPARQL endpoint. The rules that represent the English morphology are in a catalog; a forward chaining inference process is performed on top of the catalog during the extraction and linking tasks.  
\texttt{Falcon 2.0} also comprises several modules that identify and link entities and relations to the Wikidata. These modules implement POS Tagging, Tokenization \& Compounding, N-Gram Tiling, Candidate List Generation, Matching \& Ranking, Query Classifier, and N-Gram Splitting and are reused from the implementation of Falcon.

\subsection{Background Knowledge}
\begin{figure}[t]
	\centering
	\includegraphics[width=1\columnwidth]{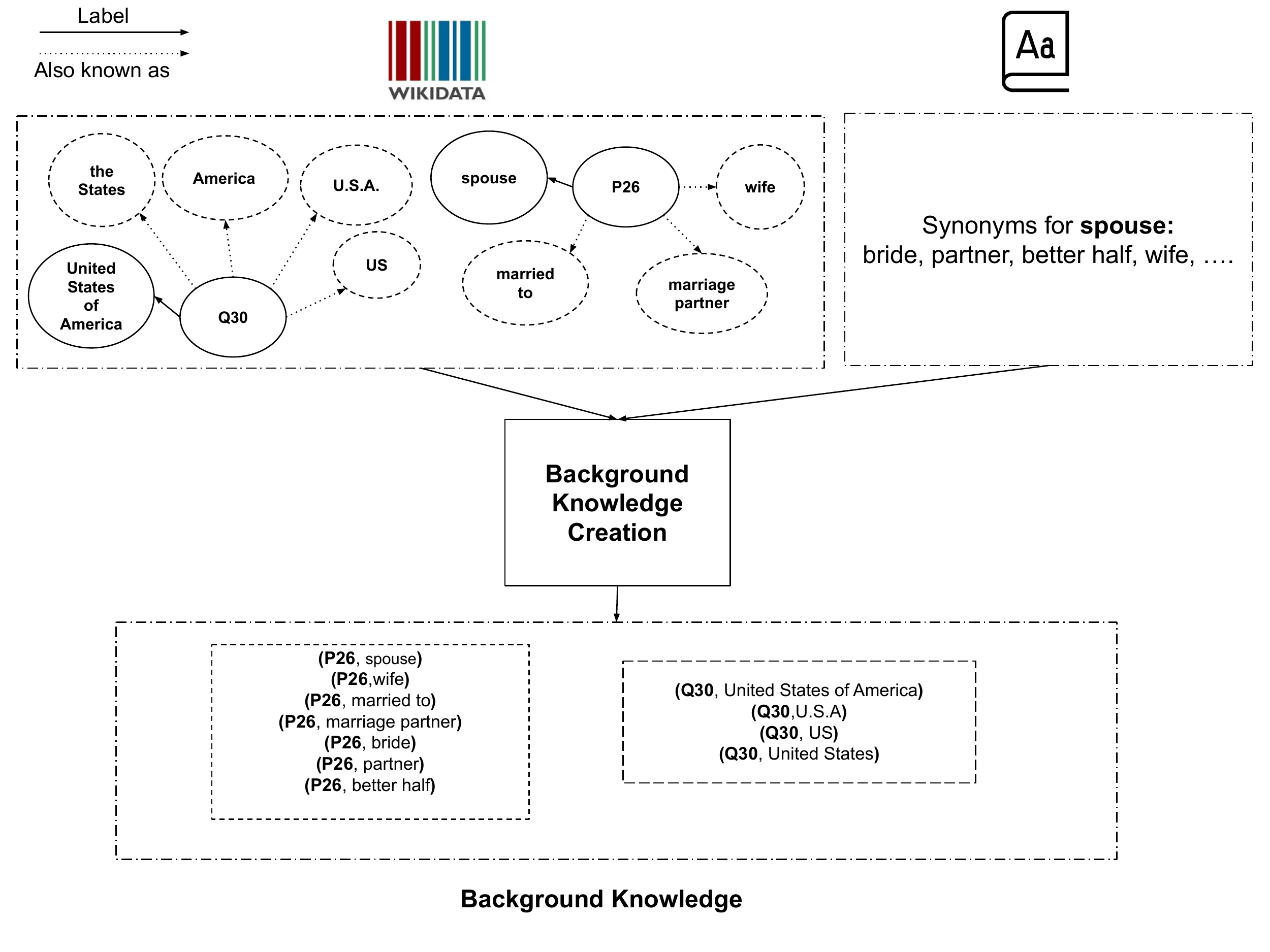}
	\caption{\textbf{\texttt{Falcon 2.0} Background Knowledge} is built by converting labels of entities and relations in Wikidata into pairs of alignments. It is a part of search engine (cf. Figure \ref{fig:archi}).
	}
	\label{fig:background}
\end{figure}
Wikidata contains over 52 million entities and 3.9 billion facts (in the form of subject-predicate-object triples). Since Falcon 2.0 background knowledge only depends on labels, a significant portion of this extensive information is not useful for our approach. Hence, we only extract all the entity and relation labels to create a local background KG, A.K.A "alias background knowledge base.". For example, the entity \texttt{United States of America}\footnote{\url{https://www.wikidata.org/wiki/Q30}} in Wikidata has the natural language label `United States of America' and several other aliases (or \texttt{known\_as} labels) of \texttt{United States of America} such as "the United States of America, America, U.S.A., the U.S., United States, etc.".
We extended our background KG with this information from Wikidata. Similarly, for relation's labels, the background KG is enriched with \texttt{known\_as} labels to provide synonyms and derived word forms. For example, the relation spouse \footnote{\url{https://www.wikidata.org/wiki/Property:P26}} in Wikidata has the label \texttt{spouse} and the other known as labels are husband, wife, married to, wedded to, partner, etc. This variety of synonyms for each relation empowers \texttt{Falcon 2.0} to match the surface form in the text to a relation in Wikidata. Figure \ref{fig:background} illustrates the process of building background knowledge.

\subsection{Catalog of Rules}
\texttt{Falcon 2.0} is a rule-based approach. A catalog of rules is predefined to extract entities and relations from the text. The rules are based on the English morphological principles and borrowed from Sakor et al.~\cite{sakor2019old}. For example, \texttt{Falcon 2.0} excludes all verbs from the entities candidates list based on the rule \texttt{verbs are not entities}. For example, the N-Gram tiling module in the \texttt{Falcon 2.0} architecture resorts to the rule: \texttt{entities with only stopwords between them are one entity}. Another example of such rule \texttt{When -> date, Where -> place} solves the ambiguity of matching the correct relation in case the short text is a question by looking at the questions headword. For example, give the two questions \texttt{When did Princess Diana die?} and \texttt{Where did Princess Diana die?}, the relation died can be the death place or the death year. The question headword (When/Where) is the only insight to solve the ambiguity here. When the question word is \texttt{where}, \texttt{Falcon 2.0} matches only relations that have a place as a range of the relation.

\subsection{Recognition}
Extraction phase in \texttt{Falcon 2.0} consists of three modules. POS tagging, tokenization \& compounding, and N-Gram tiling. The input of this phase is a natural language text. The output of the phase is the list of surface forms related to entities or relations.\\
\textbf{Part-of-speech (POS) Tagging} receives a natural language text as an input. It tags each word in the text with its related tag, e.g., noun, verb, and adverb. This module differentiates between nouns and verbs to enable the application of the morphological rules from the catalog. The output of the module is a list of the pairs of (word, tag).\\
\textbf{Tokenization \& Compounding} builds the tokens list by removing the stopwords from the input and splitting verbs from nouns. For example, if the input is \texttt{What is the operating income for Qantas}, the output of this module is a list of three tokens [operating, income, Qantas].\\
\textbf{N-Gram Tilling} module combines tokens with only stopwords between them relying on one of the rules from a catalog of rules. For example, if we consider the previous module's output as an input for the n-gram tilling module, \texttt{operating} and \texttt{income} tokens will be combined in one token. The output of the module is a list of two tokens [operating income, Qantas].
\subsection{Linking}
This phase consists of four modules: candidate list generation, matching \& ranking, relevant rule selection, and n-gram splitting. \\
\textbf{Candidate List Generation} receives the output of the recognition phase. The module queries the text search engine for each token. Then, tokens will have an associated candidate list of resources. For example, the retrieved candidate list of the token \texttt{operating income} is [(P3362, operating income), (P2139, income), (P3362, operating profit)]; where the first element is the Wikidata predicate identifier and the second is the list of labels associated with the predicates which match the query "operating income." \\
\textbf{Matching \& Ranking} ranks the candidate list received from the candidate list generation module and matches candidates' entities and relations. Since, in any KG, the facts are represented as triples, the matching and ranking module creates triples consisting of the entities and relationships from the candidates' list. Then, for each pair of entity and relation, the module checks if the triple exists in the RDF triple store (Wikidata). The checking is done by executing a simple ASK query over the RDF triple store. For each triple, the module increases the rank of the involved relations and entities. The output of the module is the ranked list of the candidates.\\
\textbf{Relevant Rule Selection} interacts with the matching \& ranking module by suggesting increasing the ranks of some candidates relying on the catalog of rules. One of the suggestions is considering the question headword to clear the ambiguity between two relations based on the range of relationships in the KG.\\
\textbf{N-Gram Splitting}  module is used if none of the triples tested in the matching \& ranking modules exists in the triple store, i.e., the compounding the approach did in the tokenization \& compounding module led to combining two separated entities. The module splits the tokens from the right side and passes the tokens again to the candidate list generation module. Splitting the tokens from the right side resorts to one of the fundamentals of the English morphology; the compound words in English have their headword always towards the right side~\cite{williams1981notions}.\\
\textbf{Text Search Engine} stores all the alignments of the labels. A simple querying technique~\cite{gormley2015elasticsearch} is used as the text search engine over background knowledge. It receives a token as an input and then returns all the related resources with labels similar to the received token. \\
\textbf{RDF Triple store} is a local copy of the Wikidata endpoint. It consists of all the RDF triples of Wikidata labeled with the English language. An RDF triple store is used to check the existence of the triples passed from the \textbf{Matching \& Ranking} module. The RDF triple store keeps around 3.9 billion triples.

\section{Experimental Study}\label{sec:experiment}
We study three research questions: RQ1) What is the performance of \texttt{Falcon 2.0} for entity linking over Wikidata? RQ2) What is the impact of Wikidata's specific background knowledge on the performance of a linguistic approach? RQ3) What is the performance of \texttt{Falcon 2.0} for relation linking over Wikidata?
\paragraph{Metrics}
We report the performance using the standard metrics of
Precision, Recall, and F-measure. Precision is the fraction of relevant resources among the retrieved resources.
Recall is the fraction of relevant resources that have been retrieved over the total amount of relevant resources.
F-Measure or F-Score is the harmonic mean of precision and recall. 
\paragraph{Datasets}
We rely on three different question answering datasets namely SimpleQuestion dataset for Wikidata~\cite{diefenbach2017question}, WebQSP-WD~\cite{sorokin2018mixing} and LC-QuAD 2.0~\cite{dubey2019lc}. The SimpleQuestion dataset contains 5,622 test questions which are answerable using Wikidata as underlying KG. WebQSP-WD contains 1639 test questions, and LC-QUAD 2.0 contains 6046 test questions. SimpleQuestion and LC-QuaD 2.0 provide the annotated gold standard for entity and relations, whereas WebQSP-WD only provides annotated gold standard for entities. Hence, we evaluated entity linking performance on three datasets and relation linking performance on two datasets. Also, SimpleQuestion and WebQSP-WD contain questions with a single entity and relation, whereas LC-QuAD 2.0 contains mostly complex questions (i.e., more than one entity and relation).
\paragraph{Baselines}
\textbf{OpenTapioca}~\cite{delpeuch2019opentapioca}: is available as a web API; it provides Wikidata URIs for entities. We run OpenTapioca API on all the three datasets.\\
\textbf{Variable Context Granularity model (VCG)}~\cite{sorokin2018mixing}: is a unifying network that
models contexts of variable granularity to extract features for mention detection and entity disambiguation. We were unable to reproduce VCG using the publicly available source code. Hence, we only report its performance on WebQSP-WD from the original paper~\cite{sorokin2018mixing} as we are unable to run the model on the other two datasets for entity linking. For the completion of the approach, we also report the other two baselines provided by the authors, namely \textbf{Heuristic Baseline} and \textbf{Simplified VCG}. \\
\textbf{S-Mart}~\cite{yang2015s}: was initially proposed to link entities in the tweets and later adapted for question answering. The system is not open source, and we adapt its result from \cite{sorokin2018mixing} for WebQSP-WD dataset. \\
\textbf{No Baseline for Relation Linking}: To the best of our knowledge, there is no baseline for relation linking on Wikidata. One argument could be to run the existing DBpedia based relation linking tool on Wikidata and compare it with our performance. We contest this solely because Wikidata is extremely noisy. For example, in "What is the longest National Highway in the world?" the entity surface form "National Highway" matches four(4) different entities in Wikidata that share the same entity label (i.e., "National Highway"). In comparison, 2,055 other entities contain the full mention in their labels for the surface form "National Highway". However, in DBpedia, there exists only one unique label for "National Highway". Hence, any entity linking tool or relation linking tool tailored for DBpedia will face issues on Wikidata (cf. table \ref{tab:falcon_webqsp}). Therefore, instead of reporting the bias and under-performance, we did not evaluate their performance for a fair comparison. Hence, we report \texttt{Falcon 2.0} relation linking performance only to establish new baselines on two datasets: SimpleQuestion and LC-QuAD 2.0. 
\begin{figure}[t]
	\centering
	\includegraphics[width=\columnwidth]{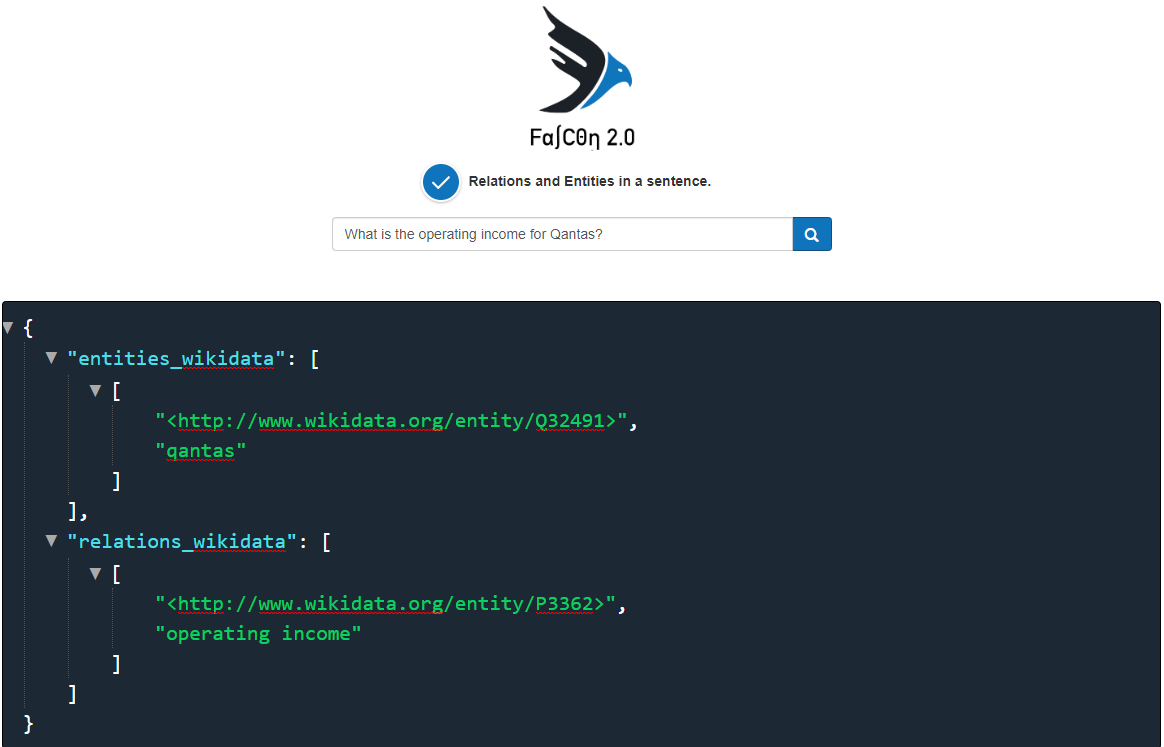}
	\caption{\textbf{\texttt{Falcon 2.0} API Web Interface}. 
	}
	\label{fig:Falcon}
\end{figure}

\paragraph{Experimental Details}
\texttt{Falcon 2.0} is extremely lightweight from an implementation point of view. A laptop machine, with eight cores and 16GB RAM running Ubuntu 18.04 is used for implementing and evaluating \texttt{Falcon 2.0}. We deployed its web API on a server with 723GB RAM, 96 cores (Intel(R) Xeon(R) Platinum 8160CPU with 2.10GHz) running Ubuntu 18.04. This publicly available API is used to calculate the standard evaluation metrics, namely Precision, Recall, and F-score.

\subsection{Experimental Results}
 \begin{table}[!t]
	\centering
	\caption{Entity linking evaluation results on LC-QuAD 2.0 \& SimpleQuestion datasets. Best values are in bold.}
	\resizebox{0.90\columnwidth}{!}{
      \begin{tabular}{ l l l l l }
   	    \toprule
         \textbf{Approach} & \textbf{Dataset} & \textbf{P} & \textbf{R} & \textbf{F} \\
               \midrule  
             {OpenTapioca \cite{delpeuch2019opentapioca}} 
               & LC-QuAD 2.0 & 0.29 & 0.42 & 0.35\\ 
            {\texttt{Falcon 2.0}}
               & LC-QuAD 2.0 & \textbf{0.50} & \textbf{0.56} & \textbf{0.53}\\ 
               {OpenTapioca \cite{delpeuch2019opentapioca}} 
               & SimpleQuestion & 0.01 & 0.02 & 0.01\\ 
               {\texttt{Falcon 2.0}}
               & SimpleQuestion  & \textbf{0.56} & \textbf{0.64} & \textbf{0.60}\\ 
               {OpenTapioca \cite{delpeuch2019opentapioca}} 
               & SimpleQuestion Uppercase Entities & 0.16 & 0.28 & 0.20\\ 
               {\texttt{Falcon 2.0}}
               & SimpleQuestion Uppercase Entities & \textbf{0.66} & \textbf{0.75} & \textbf{0.70}\\ 
            \bottomrule 
        \end{tabular}
        }
    \label{tab:Rl}
\end{table} 

\begin{table}[!t]
\centering
	\caption{Entity linking evaluation results on the WEBQSP test dataset. Best values are in bold. }
	\resizebox{0.60\columnwidth}{!}{
\begin{tabular}{llll}
\toprule
\textbf{Approach}  & \textbf{P}    & \textbf{R}    & \textbf{F}    \\ \midrule
S-MART \cite{yang2015s}        & 0.66          & 0.77          & 0.72          \\ 
Heuristic baseline \cite{sorokin2018mixing} & 0.30          & 0.61          & 0.40          \\ 
Simplified VCG \cite{sorokin2018mixing}    & \textbf{0.84} & 0.62          & 0.71          \\ 
VCG \cite{sorokin2018mixing}              & 0.83          & 0.65          & 0.73          \\ 
OpenTapioca \cite{delpeuch2019opentapioca}             & 0.01          & 0.02          & 0.02         \\ 
\texttt{Falcon 2.0}       & 0.80          & \textbf{0.84} & \textbf{0.82} \\ 
\bottomrule

\end{tabular}
}
    \label{tab:entity_webqsp}
\end{table}

\paragraph{Experimental Results 1}
In the first experiment described in Table \ref{tab:Rl}, we compare entity linking performance of \texttt{Falcon 2.0} on SimpleQuestion and LC-QuAD 2.0  datasets. We first evaluate the performance on the SimpleQuestion dataset. Surprisingly, we observe that for the OpenTapioca baseline, the values are approximately 0.0 for Precision, Recall, and F-score. We analyzed the source of errors and found that out of 5,622 questions, only 246 have entity labels in uppercase letters. Opentapioca fails to recognize and link entity mentions written in lowercase letters. Case sensitivity is a common issue for entity linking tools over short text, as reported by Singh et al.~\cite{singh2018no,DBLP:conf/www/SinghRBSLUVKP0V18} in a detailed analysis. From the remaining 246 questions, only 70 are answered correctly by OpenTapioca. Given that OpenTapioca finds limitation in linking lowercase entity surface forms. We evaluated \texttt{Falcon 2.0} and OpenTapioca on the 246 questions of SimpleQuestion to provide a fair evaluation for the baseline (reported as SimpleQuestion uppercase entities in table \ref{tab:Rl}). OpenTapioca reports F-score 0.20 on this subset of SimpleQuestion. On the other hand, \texttt{Falcon 2.0} reports F-score 0.70 on the same dataset (cf. Table \ref{tab:Rl}). For LC-QuAD 2.0, OpenTapioca reports F-score 0.35 against \texttt{Falcon 2.0} with F-score 0.53 reported in Table \ref{tab:Rl}.

\paragraph{Experimental Results 2}
We report performance of \texttt{Falcon 2.0} on WebQSP-WD dataset in Table~\ref{tab:entity_webqsp}. \texttt{Falcon 2.0} clearly outperforms all other baselines with highest F-score value 0.82. OpenTapioca demonstrates a low performance on this dataset as well. Experiment results 1 \& 2 answer our first research question (RQ1).

\paragraph{Ablation Study for Entity Linking and Recommendations}
For the second research question (RQ2), we evaluate the impact of Wikidata's specific background knowledge on the entity linking performance. 
We evaluated Falcon on the WebQSP-WD dataset against \texttt{Falcon 2.0}. We linked Falcon predicted DBpedia IRIs with corresponding Wikidata IDs using owl:sameAs. We can see in the Table~\ref{tab:falcon_webqsp} that \texttt{Falcon 2.0} significantly outperforms Falcon despite using the same linguistic driven approach. The jump in \texttt{Falcon 2.0} performance comes from Wikidata's specific local background knowledge, which we created by expanding Wikidata entities and relations with associated aliases. It also validates the novelty of \texttt{Falcon 2.0} when compared to Falcon for the Wikidata entity linking.

We observe an indifferent phenomenon in our performance for three datasets, and the performance for \texttt{Falcon 2.0} differs a lot per dataset. For instance, on WebQSP-WD, our F-score is 0.82, whereas, on LC-QuAD 2.0, the F-Score drops to 0.57. The first source of error is the dataset(s) itself. In both the datasets (SimpleQuestion and LC-QuAD 2.0), many questions are grammatically incorrect. To validate our claim more robustly, we asked two native English speakers to check the grammar of 200 random questions on LC-QuAD 2.0. Annotators reported that 42 out of 200 questions are grammatically incorrect. Many questions have erroneous spellings of the entity names. For example, "Who is the country for head of state of Mahmoud Abbas?" and "Tell me about position held of Malcolm Fraser and elected in?" are two grammatically incorrect questions in LC-QuAD 2.0.
Similarly, many questions in the SimpleQuestion dataset are also grammatically incorrect. "where was hank cochran birthed" is one such example in the SimpleQuestion dataset. \texttt{Falcon 2.0} resorts to fundamental principles of the English morphology and finds limitation in recognizing entities in many \underline{grammatically incorrect} questions. 

We also recognize that the performance of \texttt{Falcon 2.0} on sentences with minimal context is limited. For example, in the question "when did annie open?" from the WebQSP-WD dataset, the sentential context is shallow. Also, more than one instance of "Annie" exists in Wikidata, such as Wiki:Q566892 (correct one) and Wiki:Q181734. \texttt{Falcon 2.0} wrongly predicts the entity in this case.
In another example, "which country is lamb from?", the correct entity is Wiki:Q6481017 with label "lamb" in Wikidata. However, \texttt{Falcon 2.0} returns Wiki:13553878, which also has a label "lamb". In such cases, additional knowledge graph context shall prove to be useful. Approaches such as~\cite{DBLP:conf/emnlp/YangGLTZWCHR19} introduced a concept of feeding "entity descriptions" as an additional context in an entity linking model over Wikipedia. Suppose the extra context in the form of entity description (1985 English drama film directed by Colin Gregg) for the entity Wiki:13553878 is provided. In that case, a model may correctly predict the correct entity "lamb."
Based on our observations, we propose the following recommendations for the community to improve the entity linking task over Wikidata:
\begin{itemize}
  \item Wikidata has inherited challenges of vandalism and noisy entities due to crowd-authored entities~\cite{heindorf2016vandalism}. We expect the research community to come up with more robust short text datasets for the Wikidata entity linking without spelling and grammatical errors. 
  \item Rule-based approaches come with its limitations when the sentential context is minimal. However, such methods are beneficial for the nonavailability of training data. We recommend a two-step process to target questions with minimal sentential context: 1) work towards a clean and large Wikidata dataset for entity linking of short text. This will allow more robust machine learning approaches to evolve 2) use of entity descriptions from knowledge graphs to improve the linking process (same as~\cite{DBLP:conf/emnlp/YangGLTZWCHR19}).
\end{itemize}

\begin{table}[!t]
\centering
	\caption{Entity Linking Performance of Falcon vs Falcon 2.0 on WEBQSP-WD. Best values are in bold.}
	\resizebox{0.50\columnwidth}{!}{
\begin{tabular}{llll}
\toprule
\textbf{Approach}  & \textbf{P}    & \textbf{R}    & \textbf{F}    \\ \midrule
Falcon \cite{sakor2019old}           & 0.47      & 0.45          & 0.46         \\ 
\texttt{Falcon 2.0}       & \textbf{0.80 }         & \textbf{0.84} & \textbf{0.82} \\ 
\bottomrule

\end{tabular}
}
    \label{tab:falcon_webqsp}
\end{table}

 \begin{table}[!t]
	\centering
	\caption{Relation linking evaluation results on LC-QuAD 2.0 \& SimpleQuestion datasets.}
	\resizebox{0.70\columnwidth}{!}{
      \begin{tabular}{ l l l l l }
   	    \toprule
         \textbf{Approach} & \textbf{Dataset} & \textbf{P} & \textbf{R} & \textbf{F} \\
               \midrule  
            {\texttt{Falcon 2.0}}
               & LC-QuAD 2.0 & \textbf{0.44} & \textbf{0.37} & \textbf{0.40}\\ 
               {\texttt{Falcon 2.0}}
               & SimpleQuestion & \textbf{0.35} & \textbf{0.44} & \textbf{0.39}\\ 
            \bottomrule
        \end{tabular}
        }
    \label{tab:R2}
\end{table}

\begin{table*}[!t]
\centering
\caption{\textbf{Sample Questions from LC-QuAD 2.0 datset}. The table shows five sample questions and associated gold standard relations. These sentences do not include standard sentential relations in the English language. Considering Wikidata is largely authored by the crowd, the crowd often creates such uncommon relations. \texttt{Falcon 2.0} finds limitation in linking such relations, and most results are empty.}
\resizebox{\textwidth}{!}{
\begin{tabular}{ccccc}
\toprule
\textbf{Question}                                                 & \textbf{Gold Standard IDs} & \textbf{Gold Standard Labels}                        & \textbf{Predicted IDs} & \textbf{Predicted Labels} \\ \hline
Which is the global-warming potential of dichlorodifluoromethane? & P2565                      & global warming potential                             & {[}{]}                        &             \_                     \\ \hline
What is the AMCA Radiocommunications Licence ID for Qantas?       & P2472                      & ACMA Radiocommunications Client Number               & P275                          & copyright license                \\ \hline
What is ITIS TSN for Sphyraena?                                   & P815                       & ITIS TSN                                             & {[}{]}                        &       \_                           \\ \hline
What is the ARICNS for Fomalhaut?                                 & P999                       & ARICNS                                               & {[}{]}                        &                \_                  \\ \hline
Which is CIQUAL 2017 ID for cheddar?                                         & P4696          & 	
CIQUAL2017 ID &  []                          &          \_       \\ 
\bottomrule
\end{tabular}
}

\label{relationexample}
\end{table*}


\paragraph{Experimental Results 3:}
In the third experiment (for RQ3), we evaluate the relation linking performance of \texttt{Falcon 2.0}. We are not aware of any other model for relation linking over Wikidata. Table~\ref{tab:R2} summarizes relation linking performance. With this, we established new baselines over two datasets for relation linking on Wikidata. 

\paragraph{Ablation Study for Relation Linking and Recommendations}
 Falcon reported an F-score of 0.43 on LC-QuAD over DBpedia in \cite{sakor2019old} whereas \texttt{Falcon 2.0} reports a comparable relation linking F-score 0.40 on LC-QuAD 2.0 for Wikidata (cf. Table~\ref{tab:R2}). The wrong identification of the entities does affect the relation linking performance, and it is the major source of error in our case for relation linking. Table~\ref{relationexample} summarizes a sample case study for relation linking on five LC-QuAD 2.0 questions. We observe that the relations present in the questions are highly uncommon and nonstandard, and it is a peculiar property of Wikidata. \texttt{Falcon 2.0} finds limitations in linking such relations. We recommend the following:

\begin{itemize}
  \item Wikidata challenges relation linking approaches by posing a new challenge: user-created nonstandard relations such as in Table \ref{relationexample}. A rule-based approach like ours faces a clear limitation in linking such relations. Linking user-created relations in crowd-authored Wikidata is an open question for the research community. 
\end{itemize}

\section{Impact}\label{sec:impact}
In August 2019, Wikidata became the first Wikimedia project that crossed 1 billion edits, and over 20,000 active Wikidata editors\footnote{\url{https://www.wikidata.org/wiki/Wikidata:Statistics}}. A large subset of the information extraction community has extensively relied on its research around DBpedia and Wikidata targeting different research problems such as KG completion, question answering, entity linking, and data quality assessments \cite{moon2017learning,DBLP:conf/cikm/ReinandaMR16,oaqa}. Furthermore, entity and relation linking tasks have been studied well beyond information extraction research, especially NLP and Semantic Web. Despite Wikidata being hugely popular, there are limited resources for reusing and aligning unstructured text to Wikidata mentions. However, when it comes to a short text, the performance of existing baselines are limited. We believe the availability of \texttt{Falcon 2.0} as a web API along with open source access to its code will provide researchers an easy and reusable way to annotate unstructured text against Wikidata. We also believe that a rule-based approach, such as ours that does not require any training data, is beneficial for low resource languages (considering Wikidata is multilingual \footnote{\url{https://www.wikidata.org/wiki/Help:Wikimedia_language_codes/lists/all}}).


\section{Adoption and Reusability}\label{sec:adaptation}

\texttt{Falcon 2.0} is open source. The source code is available in our public GitHub: \url{https://github.com/SDM-TIB/Falcon2.0} for reusability and reproducibility. \texttt{Falcon 2.0} is easily accessible via a simple CURL request or using our web interface. Detailed instructions are provided on our GitHub. It is currently available for the English language. However, there is no assumption in the approach or while building the background knowledge base that restricts its adaptation or extensibility to other languages. The background knowledge of \texttt{Falcon 2.0} is available for the community and can be easily reused to generate candidates for entity linking~\cite{YamadaS0T16} or in question answering approaches such as~\cite{DBLP:conf/sigmod/Zhang018}. The background knowledge consists of 48,042,867 alignments of Wikidata entities and 15,645 alignments for Wikidata predicates. MIT License allows for the free distribution and re-usage of \texttt{Falcon 2.0}. We hope the research community and industry practitioners will use \texttt{Falcon 2.0} resources for various usages such as linking entities and relations to Wikidata, annotating an unstructured text, developing new low language resources, and others.

\section{Maintenance and Sustainability} \label{sec:maintance}
\texttt{Falcon 2.0} is a publicly available resource offering of the Scientific Data Management(SDM) group at TIB, Hannover\footnote{https://www.tib.eu/en/research-development/scientific-data-management/}. TIB is one of the largest libraries for Science and Technology in the world \footnote{https://www.tib.eu/en/tib/profile/}. It actively promotes open access to scientific artifacts, e.g., research data, scientific literature, non-textual material, and software. Similar to other publicly maintained repositories of SDM, \texttt{Falcon 2.0} will be preserved and regularly updated to fix bugs and include new features\footnote{\url{https://github.com/SDM-TIB}}. The \texttt{Falcon 2.0} API will be sustained on the TIB servers to allow for unrestricted free access.

\section{Conclusion and Future Work} \label{sec:conclude}
We presented the resource \texttt{Falcon 2.0}, a rule-based entity and relation linking tool able to recognize entities \& relations in a short text and link them to the existing knowledge graph, e.g., DBpedia and Wikidata. Although there are various approaches for entity \& relation linking to DBpedia, \texttt{Falcon 2.0} is one of the few tools targeting Wikidata. Thus, given the number of generic and domain-specific facts that compose Wikidata, \texttt{Falcon 2.0} has the potential to impact researchers and practitioners that resort to NLP tools for transforming semi-structured data into structured facts. \texttt{Falcon 2.0} is open source. The API is publicly accessible and maintained in the servers of the TIB labs. \texttt{Falcon 2.0} has been empirically evaluated on three benchmarks, and the outcomes suggest that it is able to overcome the state of the arts. Albeit promising, the experimental results can be improved. In the future, we plan to continue researching novel techniques that enable adjusting the catalog of rules and alignments to the changes in Wikidata. We further plan to mitigate errors caused by a rule-based approach using machine learning approaches to aim towards a hybrid approach.

\section{Acknowledgments}
This work has received funding from the EU H2020 Project No. 727658 (IASIS).

\bibliographystyle{ACM-Reference-Format}
\bibliography{sample-sigconf}


\begin{thebibliography}{35}


\ifx \showCODEN    \undefined \def \showCODEN     #1{\unskip}     \fi
\ifx \showDOI      \undefined \def \showDOI       #1{#1}\fi
\ifx \showISBNx    \undefined \def \showISBNx     #1{\unskip}     \fi
\ifx \showISBNxiii \undefined \def \showISBNxiii  #1{\unskip}     \fi
\ifx \showISSN     \undefined \def \showISSN      #1{\unskip}     \fi
\ifx \showLCCN     \undefined \def \showLCCN      #1{\unskip}     \fi
\ifx \shownote     \undefined \def \shownote      #1{#1}          \fi
\ifx \showarticletitle \undefined \def \showarticletitle #1{#1}   \fi
\ifx \showURL      \undefined \def \showURL       {\relax}        \fi
\providecommand\bibfield[2]{#2}
\providecommand\bibinfo[2]{#2}
\providecommand\natexlab[1]{#1}
\providecommand\showeprint[2][]{arXiv:#2}

\bibitem[\protect\citeauthoryear{Auer, Bizer, Kobilarov, Lehmann, Cyganiak, and
  Ives}{Auer et~al\mbox{.}}{2007}]%
        {DBLP:conf/semweb/AuerBKLCI07}
\bibfield{author}{\bibinfo{person}{S{\"{o}}ren Auer},
  \bibinfo{person}{Christian Bizer}, \bibinfo{person}{Georgi Kobilarov},
  \bibinfo{person}{Jens Lehmann}, \bibinfo{person}{Richard Cyganiak}, {and}
  \bibinfo{person}{Zachary~G. Ives}.} \bibinfo{year}{2007}\natexlab{}.
\newblock \showarticletitle{{DBpedia: A Nucleus for a Web of Open Data}}. In
  \bibinfo{booktitle}{\emph{{ISWC}}}. \bibinfo{pages}{722--735}.
\newblock


\bibitem[\protect\citeauthoryear{Balog}{Balog}{2018}]%
        {balog_2018}
\bibfield{author}{\bibinfo{person}{Krisztian Balog}.}
  \bibinfo{year}{2018}\natexlab{}.
\newblock \bibinfo{booktitle}{\emph{Entity-oriented search}}.
\newblock \bibinfo{publisher}{Springer Open}.
\newblock


\bibitem[\protect\citeauthoryear{Banerjee, Dubey, Chaudhuri, and
  Lehmann}{Banerjee et~al\mbox{.}}{[n.d.]}]%
        {banerjeejoint}
\bibfield{author}{\bibinfo{person}{Debayan Banerjee}, \bibinfo{person}{Mohnish
  Dubey}, \bibinfo{person}{Debanjan Chaudhuri}, {and} \bibinfo{person}{Jens
  Lehmann}.} \bibinfo{year}{[n.d.]}\natexlab{}.
\newblock \showarticletitle{Joint Entity and Relation Linking using EARL}.
\newblock  (\bibinfo{year}{[n.\,d.]}).
\newblock


\bibitem[\protect\citeauthoryear{Bollacker, Evans, Paritosh, Sturge, and
  Taylor}{Bollacker et~al\mbox{.}}{2008}]%
        {DBLP:conf/aaai/BollackerCT07}
\bibfield{author}{\bibinfo{person}{Kurt~D. Bollacker}, \bibinfo{person}{Colin
  Evans}, \bibinfo{person}{Praveen Paritosh}, \bibinfo{person}{Tim Sturge},
  {and} \bibinfo{person}{Jamie Taylor}.} \bibinfo{year}{2008}\natexlab{}.
\newblock \showarticletitle{Freebase: a collaboratively created graph database
  for structuring human knowledge}. In \bibinfo{booktitle}{\emph{{ACM}
  {SIGMOD}}}. \bibinfo{pages}{1247--1250}.
\newblock


\bibitem[\protect\citeauthoryear{Cao, Hou, Li, and Liu}{Cao
  et~al\mbox{.}}{2018}]%
        {CaoYixin-2018}
\bibfield{author}{\bibinfo{person}{Yixin Cao}, \bibinfo{person}{Lei Hou},
  \bibinfo{person}{Juanzi Li}, {and} \bibinfo{person}{Zhiyuan Liu}.}
  \bibinfo{year}{2018}\natexlab{}.
\newblock \bibinfo{title}{Neural Collective Entity Linking}.
\newblock
\newblock
\showeprint[arxiv]{1811.08603}
\urldef\tempurl%
\url{http://arxiv.org/abs/1811.08603}
\showURL{%
\tempurl}


\bibitem[\protect\citeauthoryear{Cetoli, Bragaglia, O’Harney, Sloan, and
  Akbari}{Cetoli et~al\mbox{.}}{2019}]%
        {cetoli2019neural}
\bibfield{author}{\bibinfo{person}{Alberto Cetoli}, \bibinfo{person}{Stefano
  Bragaglia}, \bibinfo{person}{Andrew~D O’Harney}, \bibinfo{person}{Marc
  Sloan}, {and} \bibinfo{person}{Mohammad Akbari}.}
  \bibinfo{year}{2019}\natexlab{}.
\newblock \showarticletitle{A Neural Approach to Entity Linking on Wikidata}.
  In \bibinfo{booktitle}{\emph{European Conference on Information Retrieval}}.
  Springer, \bibinfo{pages}{78--86}.
\newblock


\bibitem[\protect\citeauthoryear{Delpeuch}{Delpeuch}{2019}]%
        {delpeuch2019opentapioca}
\bibfield{author}{\bibinfo{person}{Antonin Delpeuch}.}
  \bibinfo{year}{2019}\natexlab{}.
\newblock \showarticletitle{OpenTapioca: Lightweight Entity Linking for
  Wikidata}.
\newblock \bibinfo{journal}{\emph{arXiv preprint arXiv:1904.09131}}
  (\bibinfo{year}{2019}).
\newblock


\bibitem[\protect\citeauthoryear{Diefenbach, Tanon, Singh, and
  Maret}{Diefenbach et~al\mbox{.}}{2017}]%
        {diefenbach2017question}
\bibfield{author}{\bibinfo{person}{Dennis Diefenbach}, \bibinfo{person}{Thomas
  Tanon}, \bibinfo{person}{Kamal Singh}, {and} \bibinfo{person}{Pierre Maret}.}
  \bibinfo{year}{2017}\natexlab{}.
\newblock \showarticletitle{Question answering benchmarks for wikidata}.
\newblock


\bibitem[\protect\citeauthoryear{Dubey, Banerjee, Abdelkawi, and Lehmann}{Dubey
  et~al\mbox{.}}{2019}]%
        {dubey2019lc}
\bibfield{author}{\bibinfo{person}{Mohnish Dubey}, \bibinfo{person}{Debayan
  Banerjee}, \bibinfo{person}{Abdelrahman Abdelkawi}, {and}
  \bibinfo{person}{Jens Lehmann}.} \bibinfo{year}{2019}\natexlab{}.
\newblock \showarticletitle{Lc-quad 2.0: A large dataset for complex question
  answering over wikidata and dbpedia}. In
  \bibinfo{booktitle}{\emph{International Semantic Web Conference}}. Springer,
  \bibinfo{pages}{69--78}.
\newblock


\bibitem[\protect\citeauthoryear{Ferragina and Scaiella}{Ferragina and
  Scaiella}{2010}]%
        {DBLP:conf/cikm/FerraginaS10}
\bibfield{author}{\bibinfo{person}{Paolo Ferragina} {and} \bibinfo{person}{Ugo
  Scaiella}.} \bibinfo{year}{2010}\natexlab{}.
\newblock \showarticletitle{{TAGME:} on-the-fly annotation of short text
  fragments (by wikipedia entities)}. In \bibinfo{booktitle}{\emph{{Proceedings
  of the 19th ACM Conference on Information and Knowledge Management, CIKM
  2010, Toronto, Ontario, Canada, October 26-30, 2010}}}.
  \bibinfo{pages}{1625--1628}.
\newblock


\bibitem[\protect\citeauthoryear{Ganea and Hofmann}{Ganea and Hofmann}{2017}]%
        {DBLP:conf/emnlp/GaneaH17}
\bibfield{author}{\bibinfo{person}{Octavian{-}Eugen Ganea} {and}
  \bibinfo{person}{Thomas Hofmann}.} \bibinfo{year}{2017}\natexlab{}.
\newblock \showarticletitle{Deep Joint Entity Disambiguation with Local Neural
  Attention}. In \bibinfo{booktitle}{\emph{Proceedings of the 2017 Conference
  on Empirical Methods in Natural Language Processing, {EMNLP} 2017,
  Copenhagen, Denmark, September 9-11, 2017}}. \bibinfo{pages}{2619--2629}.
\newblock


\bibitem[\protect\citeauthoryear{Gormley and Tong}{Gormley and Tong}{2015}]%
        {gormley2015elasticsearch}
\bibfield{author}{\bibinfo{person}{Clinton Gormley} {and}
  \bibinfo{person}{Zachary Tong}.} \bibinfo{year}{2015}\natexlab{}.
\newblock \bibinfo{booktitle}{\emph{Elasticsearch: The Definitive Guide: A
  Distributed Real-Time Search and Analytics Engine}}.
\newblock \bibinfo{publisher}{" O'Reilly Media, Inc."}.
\newblock


\bibitem[\protect\citeauthoryear{Heindorf, Potthast, Stein, and
  Engels}{Heindorf et~al\mbox{.}}{2016}]%
        {heindorf2016vandalism}
\bibfield{author}{\bibinfo{person}{Stefan Heindorf}, \bibinfo{person}{Martin
  Potthast}, \bibinfo{person}{Benno Stein}, {and} \bibinfo{person}{Gregor
  Engels}.} \bibinfo{year}{2016}\natexlab{}.
\newblock \showarticletitle{Vandalism detection in wikidata}. In
  \bibinfo{booktitle}{\emph{Proceedings of the 25th ACM International on
  Conference on Information and Knowledge Management}}.
  \bibinfo{pages}{327--336}.
\newblock


\bibitem[\protect\citeauthoryear{Hoffart, Yosef, Bordino, F{\"{u}}rstenau,
  Pinkal, Spaniol, Taneva, Thater, and Weikum}{Hoffart et~al\mbox{.}}{2011}]%
        {DBLP:conf/emnlp/HoffartYBFPSTTW11}
\bibfield{author}{\bibinfo{person}{Johannes Hoffart},
  \bibinfo{person}{Mohamed~Amir Yosef}, \bibinfo{person}{Ilaria Bordino},
  \bibinfo{person}{Hagen F{\"{u}}rstenau}, \bibinfo{person}{Manfred Pinkal},
  \bibinfo{person}{Marc Spaniol}, \bibinfo{person}{Bilyana Taneva},
  \bibinfo{person}{Stefan Thater}, {and} \bibinfo{person}{Gerhard Weikum}.}
  \bibinfo{year}{2011}\natexlab{}.
\newblock \showarticletitle{{Robust Disambiguation of Named Entities in Text}}.
  In \bibinfo{booktitle}{\emph{{EMNLP 2011}}}. \bibinfo{pages}{782--792}.
\newblock


\bibitem[\protect\citeauthoryear{Inan and Dikenelli}{Inan and
  Dikenelli}{2018}]%
        {Emrah-W18-2403}
\bibfield{author}{\bibinfo{person}{Emrah Inan} {and} \bibinfo{person}{Oguz
  Dikenelli}.} \bibinfo{year}{2018}\natexlab{}.
\newblock \showarticletitle{A Sequence Learning Method for Domain-Specific
  Entity Linking}. In \bibinfo{booktitle}{\emph{Proceedings of the Seventh
  Named Entities Workshop}} (Melbourne, Australia).
  \bibinfo{publisher}{Association for Computational Linguistics},
  \bibinfo{pages}{14--21}.
\newblock
\urldef\tempurl%
\url{http://aclweb.org/anthology/W18-2403}
\showURL{%
\tempurl}


\bibitem[\protect\citeauthoryear{Ji}{Ji}{2019}]%
        {hengji2019}
\bibfield{author}{\bibinfo{person}{Heng Ji}.} \bibinfo{year}{2019}\natexlab{}.
\newblock \bibinfo{booktitle}{\emph{Entity Discovery and Linking and
  Wikification Reading List}}.
\newblock
\urldef\tempurl%
\url{http://nlp.cs.rpi.edu/kbp/2014/elreading.html}
\showURL{%
\tempurl}


\bibitem[\protect\citeauthoryear{Kolitsas, Ganea, and Hofmann}{Kolitsas
  et~al\mbox{.}}{2018}]%
        {kolitsas2018end}
\bibfield{author}{\bibinfo{person}{Nikolaos Kolitsas},
  \bibinfo{person}{Octavian-Eugen Ganea}, {and} \bibinfo{person}{Thomas
  Hofmann}.} \bibinfo{year}{2018}\natexlab{}.
\newblock \showarticletitle{End-to-End Neural Entity Linking}. In
  \bibinfo{booktitle}{\emph{Proceedings of the 22nd Conference on Computational
  Natural Language Learning}}. \bibinfo{pages}{519--529}.
\newblock


\bibitem[\protect\citeauthoryear{Moon, Jones, and Samatova}{Moon
  et~al\mbox{.}}{2017}]%
        {moon2017learning}
\bibfield{author}{\bibinfo{person}{Changsung Moon}, \bibinfo{person}{Paul
  Jones}, {and} \bibinfo{person}{Nagiza~F Samatova}.}
  \bibinfo{year}{2017}\natexlab{}.
\newblock \showarticletitle{Learning entity type embeddings for knowledge graph
  completion}. In \bibinfo{booktitle}{\emph{Proceedings of the 2017 ACM on
  conference on information and knowledge management}}.
  \bibinfo{pages}{2215--2218}.
\newblock


\bibitem[\protect\citeauthoryear{Mulang, Singh, Vyas, Shekarpour, Sakor, Vidal,
  Auer, and Lehmann}{Mulang et~al\mbox{.}}{2020}]%
        {mulang2019context}
\bibfield{author}{\bibinfo{person}{Isaiah~Onando Mulang},
  \bibinfo{person}{Kuldeep Singh}, \bibinfo{person}{Akhilesh Vyas},
  \bibinfo{person}{Saeedeh Shekarpour}, \bibinfo{person}{Ahmad Sakor},
  \bibinfo{person}{Maria~Esther Vidal}, \bibinfo{person}{Soren Auer}, {and}
  \bibinfo{person}{Jens Lehmann}.} \bibinfo{year}{2020}\natexlab{}.
\newblock \showarticletitle{Encoding Knowledge Graph Entity Aliases in an
  Attentive Neural Networks for Wikidata Entity Linking}.
\newblock \bibinfo{journal}{\emph{In WISE (to appear)}} (\bibinfo{year}{2020}).
\newblock


\bibitem[\protect\citeauthoryear{Raiman and Raiman}{Raiman and Raiman}{2018}]%
        {raiman2018deeptype}
\bibfield{author}{\bibinfo{person}{Jonathan~Raphael Raiman} {and}
  \bibinfo{person}{Olivier~Michel Raiman}.} \bibinfo{year}{2018}\natexlab{}.
\newblock \showarticletitle{DeepType: multilingual entity linking by neural
  type system evolution}. In \bibinfo{booktitle}{\emph{Thirty-Second AAAI
  Conference on Artificial Intelligence}}.
\newblock


\bibitem[\protect\citeauthoryear{Reinanda, Meij, and de~Rijke}{Reinanda
  et~al\mbox{.}}{2016}]%
        {DBLP:conf/cikm/ReinandaMR16}
\bibfield{author}{\bibinfo{person}{Ridho Reinanda}, \bibinfo{person}{Edgar
  Meij}, {and} \bibinfo{person}{Maarten de Rijke}.}
  \bibinfo{year}{2016}\natexlab{}.
\newblock \showarticletitle{{Document Filtering for Long-tail Entities}}. In
  \bibinfo{booktitle}{\emph{{Proceedings of the 25th ACM International
  Conference on Information and Knowledge Management, CIKM 2016, Indianapolis,
  IN, USA, October 24-28, 2016}}}. \bibinfo{publisher}{{ACM}},
  \bibinfo{pages}{771--780}.
\newblock
\urldef\tempurl%
\url{https://doi.org/10.1145/2983323.2983728}
\showDOI{\tempurl}


\bibitem[\protect\citeauthoryear{R{\"o}der, Usbeck, and Ngonga~Ngomo}{R{\"o}der
  et~al\mbox{.}}{2018}]%
        {roder2018gerbil}
\bibfield{author}{\bibinfo{person}{Michael R{\"o}der}, \bibinfo{person}{Ricardo
  Usbeck}, {and} \bibinfo{person}{Axel-Cyrille Ngonga~Ngomo}.}
  \bibinfo{year}{2018}\natexlab{}.
\newblock \showarticletitle{Gerbil--benchmarking named entity recognition and
  linking consistently}.
\newblock \bibinfo{journal}{\emph{Semantic Web}} \bibinfo{volume}{9},
  \bibinfo{number}{5} (\bibinfo{year}{2018}), \bibinfo{pages}{605--625}.
\newblock


\bibitem[\protect\citeauthoryear{Sakor, Mulang, Singh, Shekarpour, Vidal,
  Lehmann, and Auer}{Sakor et~al\mbox{.}}{2019}]%
        {sakor2019old}
\bibfield{author}{\bibinfo{person}{Ahmad Sakor}, \bibinfo{person}{Isaiah~Onando
  Mulang}, \bibinfo{person}{Kuldeep Singh}, \bibinfo{person}{Saeedeh
  Shekarpour}, \bibinfo{person}{Maria~Esther Vidal}, \bibinfo{person}{Jens
  Lehmann}, {and} \bibinfo{person}{S{\"o}ren Auer}.}
  \bibinfo{year}{2019}\natexlab{}.
\newblock \showarticletitle{Old is gold: linguistic driven approach for entity
  and relation linking of short text}. In \bibinfo{booktitle}{\emph{Proceedings
  of the 2019 NAACL HLT (Long Papers)}}. \bibinfo{pages}{2336--2346}.
\newblock


\bibitem[\protect\citeauthoryear{Shen, Wang, and Han}{Shen
  et~al\mbox{.}}{2015}]%
        {shen2015}
\bibfield{author}{\bibinfo{person}{W. Shen}, \bibinfo{person}{J. Wang}, {and}
  \bibinfo{person}{J. Han}.} \bibinfo{year}{2015}\natexlab{}.
\newblock \showarticletitle{Entity Linking with a Knowledge Base: Issues,
  Techniques, and Solutions}.
\newblock \bibinfo{journal}{\emph{IEEE Transactions on Knowledge and Data
  Engineering}} \bibinfo{volume}{27}, \bibinfo{number}{2}
  (\bibinfo{year}{2015}), \bibinfo{pages}{443--460}.
\newblock


\bibitem[\protect\citeauthoryear{Singh, Lytra, Radhakrishna, Shekarpour, Vidal,
  and Lehmann}{Singh et~al\mbox{.}}{2018a}]%
        {singh2018no}
\bibfield{author}{\bibinfo{person}{Kuldeep Singh}, \bibinfo{person}{Ioanna
  Lytra}, \bibinfo{person}{Arun~Sethupat Radhakrishna},
  \bibinfo{person}{Saeedeh Shekarpour}, \bibinfo{person}{Maria-Esther Vidal},
  {and} \bibinfo{person}{Jens Lehmann}.} \bibinfo{year}{2018}\natexlab{a}.
\newblock \showarticletitle{No One is Perfect: Analysing the Performance of
  Question Answering Components over the DBpedia Knowledge Graph}.
\newblock \bibinfo{journal}{\emph{arXiv:1809.10044}} (\bibinfo{year}{2018}).
\newblock


\bibitem[\protect\citeauthoryear{Singh, Radhakrishna, Both, Shekarpour, Lytra,
  Usbeck, Vyas, Khikmatullaev, Punjani, Lange, Vidal, Lehmann, and Auer}{Singh
  et~al\mbox{.}}{2018b}]%
        {DBLP:conf/www/SinghRBSLUVKP0V18}
\bibfield{author}{\bibinfo{person}{Kuldeep Singh},
  \bibinfo{person}{Arun~Sethupat Radhakrishna}, \bibinfo{person}{Andreas Both},
  \bibinfo{person}{Saeedeh Shekarpour}, \bibinfo{person}{Ioanna Lytra},
  \bibinfo{person}{Ricardo Usbeck}, \bibinfo{person}{Akhilesh Vyas},
  \bibinfo{person}{Akmal Khikmatullaev}, \bibinfo{person}{Dharmen Punjani},
  \bibinfo{person}{Christoph Lange}, \bibinfo{person}{Maria{-}Esther Vidal},
  \bibinfo{person}{Jens Lehmann}, {and} \bibinfo{person}{S{\"{o}}ren Auer}.}
  \bibinfo{year}{2018}\natexlab{b}.
\newblock \showarticletitle{{Why Reinvent the Wheel: Let's Build Question
  Answering Systems Together}}. In \bibinfo{booktitle}{\emph{{Web
  Conference}}}. \bibinfo{pages}{1247--1256}.
\newblock


\bibitem[\protect\citeauthoryear{Sorokin and Gurevych}{Sorokin and
  Gurevych}{2018}]%
        {sorokin2018mixing}
\bibfield{author}{\bibinfo{person}{Daniil Sorokin} {and} \bibinfo{person}{Iryna
  Gurevych}.} \bibinfo{year}{2018}\natexlab{}.
\newblock \showarticletitle{Mixing Context Granularities for Improved Entity
  Linking on Question Answering Data across Entity Categories}. In
  \bibinfo{booktitle}{\emph{Proceedings of the Seventh Joint Conference on
  Lexical and Computational Semantics}}. \bibinfo{pages}{65--75}.
\newblock


\bibitem[\protect\citeauthoryear{Vrandecic}{Vrandecic}{2012}]%
        {DBLP:conf/www/Vrandecic12}
\bibfield{author}{\bibinfo{person}{Denny Vrandecic}.}
  \bibinfo{year}{2012}\natexlab{}.
\newblock \showarticletitle{Wikidata: a new platform for collaborative data
  collection}. In \bibinfo{booktitle}{\emph{{Proceedings of the 21st World Wide
  Web Conference, {WWW} 2012, Lyon, France, April 16-20, 2012 (Companion
  Volume)}}}. \bibinfo{publisher}{{ACM}}, \bibinfo{pages}{1063--1064}.
\newblock
\urldef\tempurl%
\url{https://doi.org/10.1145/2187980.2188242}
\showDOI{\tempurl}


\bibitem[\protect\citeauthoryear{Williams}{Williams}{1981}]%
        {williams1981notions}
\bibfield{author}{\bibinfo{person}{Edwin Williams}.}
  \bibinfo{year}{1981}\natexlab{}.
\newblock \showarticletitle{On the notions" Lexically related" and" Head of a
  word"}.
\newblock \bibinfo{journal}{\emph{Linguistic inquiry}} \bibinfo{volume}{12},
  \bibinfo{number}{2} (\bibinfo{year}{1981}), \bibinfo{pages}{245--274}.
\newblock


\bibitem[\protect\citeauthoryear{Yamada, Shindo, Takeda, and Takefuji}{Yamada
  et~al\mbox{.}}{2016a}]%
        {DBLP:conf/conll/YamadaS0T16}
\bibfield{author}{\bibinfo{person}{Ikuya Yamada}, \bibinfo{person}{Hiroyuki
  Shindo}, \bibinfo{person}{Hideaki Takeda}, {and} \bibinfo{person}{Yoshiyasu
  Takefuji}.} \bibinfo{year}{2016}\natexlab{a}.
\newblock \showarticletitle{Joint Learning of the Embedding of Words and
  Entities for Named Entity Disambiguation}. In \bibinfo{booktitle}{\emph{CoNLL
  2016}}, \bibfield{editor}{\bibinfo{person}{Yoav Goldberg} {and}
  \bibinfo{person}{Stefan Riezler}} (Eds.). \bibinfo{publisher}{{ACL}},
  \bibinfo{pages}{250--259}.
\newblock


\bibitem[\protect\citeauthoryear{Yamada, Shindo, Takeda, and Takefuji}{Yamada
  et~al\mbox{.}}{2016b}]%
        {YamadaS0T16}
\bibfield{author}{\bibinfo{person}{Ikuya Yamada}, \bibinfo{person}{Hiroyuki
  Shindo}, \bibinfo{person}{Hideaki Takeda}, {and} \bibinfo{person}{Yoshiyasu
  Takefuji}.} \bibinfo{year}{2016}\natexlab{b}.
\newblock \showarticletitle{Joint Learning of the Embedding of Words and
  Entities for Named Entity Disambiguation}.
\newblock \bibinfo{journal}{\emph{CoRR}}  \bibinfo{volume}{abs/1601.01343}
  (\bibinfo{year}{2016}).
\newblock


\bibitem[\protect\citeauthoryear{Yang, Gu, Lin, Tang, Zhuang, Wu, Chen, Hu, and
  Ren}{Yang et~al\mbox{.}}{2019}]%
        {DBLP:conf/emnlp/YangGLTZWCHR19}
\bibfield{author}{\bibinfo{person}{Xiyuan Yang}, \bibinfo{person}{Xiaotao Gu},
  \bibinfo{person}{Sheng Lin}, \bibinfo{person}{Siliang Tang},
  \bibinfo{person}{Yueting Zhuang}, \bibinfo{person}{Fei Wu},
  \bibinfo{person}{Zhigang Chen}, \bibinfo{person}{Guoping Hu}, {and}
  \bibinfo{person}{Xiang Ren}.} \bibinfo{year}{2019}\natexlab{}.
\newblock \showarticletitle{Learning Dynamic Context Augmentation for Global
  Entity Linking}. In \bibinfo{booktitle}{\emph{{EMNLP-IJCNLP} 2019}},
  \bibfield{editor}{\bibinfo{person}{Kentaro Inui}, \bibinfo{person}{Jing
  Jiang}, \bibinfo{person}{Vincent Ng}, {and} \bibinfo{person}{Xiaojun Wan}}
  (Eds.). \bibinfo{pages}{271--281}.
\newblock


\bibitem[\protect\citeauthoryear{Yang and Chang}{Yang and Chang}{2015}]%
        {yang2015s}
\bibfield{author}{\bibinfo{person}{Yi Yang} {and} \bibinfo{person}{Ming-Wei
  Chang}.} \bibinfo{year}{2015}\natexlab{}.
\newblock \showarticletitle{S-MART: Novel Tree-based Structured Learning
  Algorithms Applied to Tweet Entity Linking}. In
  \bibinfo{booktitle}{\emph{ACL- IJCNLP (Volume 1: Long Papers)}}.
  \bibinfo{pages}{504--513}.
\newblock


\bibitem[\protect\citeauthoryear{Yang, Gardu{\~{n}}o, Fang, Maiberg, McCormack,
  and Nyberg}{Yang et~al\mbox{.}}{2013}]%
        {oaqa}
\bibfield{author}{\bibinfo{person}{Zi Yang}, \bibinfo{person}{Elmer
  Gardu{\~{n}}o}, \bibinfo{person}{Yan Fang}, \bibinfo{person}{Avner Maiberg},
  \bibinfo{person}{Collin McCormack}, {and} \bibinfo{person}{Eric Nyberg}.}
  \bibinfo{year}{2013}\natexlab{}.
\newblock \showarticletitle{Building optimal information systems automatically:
  configuration space exploration for biomedical information systems}. In
  \bibinfo{booktitle}{\emph{{22nd {ACM} CIKM'13, San Francisco, USA}}}.
  \bibinfo{publisher}{{ACM}}, \bibinfo{pages}{1421--1430}.
\newblock


\bibitem[\protect\citeauthoryear{Zhang and Zou}{Zhang and Zou}{2018}]%
        {DBLP:conf/sigmod/Zhang018}
\bibfield{author}{\bibinfo{person}{Xinbo Zhang} {and} \bibinfo{person}{Lei
  Zou}.} \bibinfo{year}{2018}\natexlab{}.
\newblock \showarticletitle{{IMPROVE-QA: An Interactive Mechanism for RDF
  Question/Answering Systems}}. In \bibinfo{booktitle}{\emph{{Proceedings of
  the 2018 International Conference on Management of Data, {SIGMOD} Conference
  2018, Houston, TX, USA, June 10-15, 2018}}}. \bibinfo{pages}{1753--1756}.
\newblock
\urldef\tempurl%
\url{https://doi.org/10.1145/3183713.3193555}
\showDOI{\tempurl}


\end{thebibliography}

\end{document}